\documentclass{article}
\usepackage{amsmath,epsfig}
\usepackage[preprint]{spconfa4}

\usepackage{xcolor}
\usepackage{amssymb}
\usepackage[switch]{lineno}
\usepackage{subcaption}
\usepackage{algpseudocode}
\usepackage{float}
\usepackage{hyperref}
\usepackage{enumitem}

\usepackage{xspace}

\copyrightnotice{978-1-6654-3864-3/21/\$31.00 ©2021 IEEE}

\let\OLDthebibliography\thebibliography
\renewcommand\thebibliography[1]{
  \OLDthebibliography{#1}
  \setlength{\parskip}{0pt}
  \setlength{\itemsep}{0pt plus 0.3ex}
}

\newcommand{\modelname}{\text{SAFIN}}

\begin{document}\sloppy
\topmargin=0mm
\def\x{{\mathbf x}}
\def\L{{\cal L}}

\title{SAFIN: Arbitrary Style Transfer with Self-Attentive Factorized Instance Normalization}
%
\name{
    Aaditya Singh\thanks{\textsuperscript{*} Equal contribution}\thanks{\textsuperscript{\textdagger} Currently affiliated with IBM Research}\textsuperscript{*\textdagger\rm 1},
    Shreeshail Hingane\textsuperscript{*\rm 1},
    Xinyu Gong\textsuperscript{\rm 2},
    Zhangyang Wang\textsuperscript{\rm 2}
}

\address{
    \textsuperscript{\rm 1} Indian Institute of Technology Kanpur,
    \textsuperscript{\rm 2} The University of Texas at Austin
}

\maketitle

\begin{abstract}
Artistic style transfer aims to transfer the style characteristics of one image onto another image while retaining its content. Existing approaches commonly leverage various normalization techniques, although these face limitations in adequately transferring diverse textures to different spatial locations. Self-Attention based approaches have tackled this issue with partial success but suffer from unwanted artifacts. Motivated by these observations, this paper aims to combine the best of both worlds: self-attention and normalization. That yields a new plug-and-play module that we name \textit{Self-Attentive Factorized Instance Normalization} (\textbf{\modelname}). \modelname \xspace is essentially a spatially-adaptive normalization module whose parameters are inferred through attention on the content and style image. We demonstrate that plugging \modelname \xspace into the base network of another state-of-the-art method results in enhanced stylization. We also develop a novel base network composed of Wavelet Transform for multi-scale style transfer, which when combined with \modelname, produces visually appealing results with lesser unwanted textures. 
\end{abstract}
%
%

\nocite{*}
\section{Introduction}
Artistic style transfer is the task of extracting style and texture patterns from one image and transferring them onto the content of another. Initial work by \cite{gatys2016image} employed an optimization based approach. This approach was replaced by feed-forward networks that could generate the images in a single forward pass \cite{johnson2016perceptual} \cite{ulyanov2016texture}, with the major limitation being that these had to be retrained for every new style image. Further work on doing \textit{multiple-styles-per-model} followed \cite{dumoulin2016learned}, though these could not effectively generalize to new unseen styles. 
\par Further research led to models that perform style transfer for \textit{arbitrary} styles \cite{huang2017arbitrary} \cite{li2017universal}. Adaptive Instance Normalization (AdaIN) \cite{huang2017arbitrary} was one of the enabling techniques, which adjusted the mean and variance of the content image feature-maps to match those of the style image. The transferred results, though often decent, also appear to be over-simplistic and lack fine details, since only simple statistic moments were drawn for content-style matching. Following this idea, many other normalization based approaches such as Batch-IN \cite{nam2018batch} and Dynamic-IN \cite{jing2019dynamic} were proposed.
\begin{figure}[t!]
    \centering
    \includegraphics[scale=0.305]{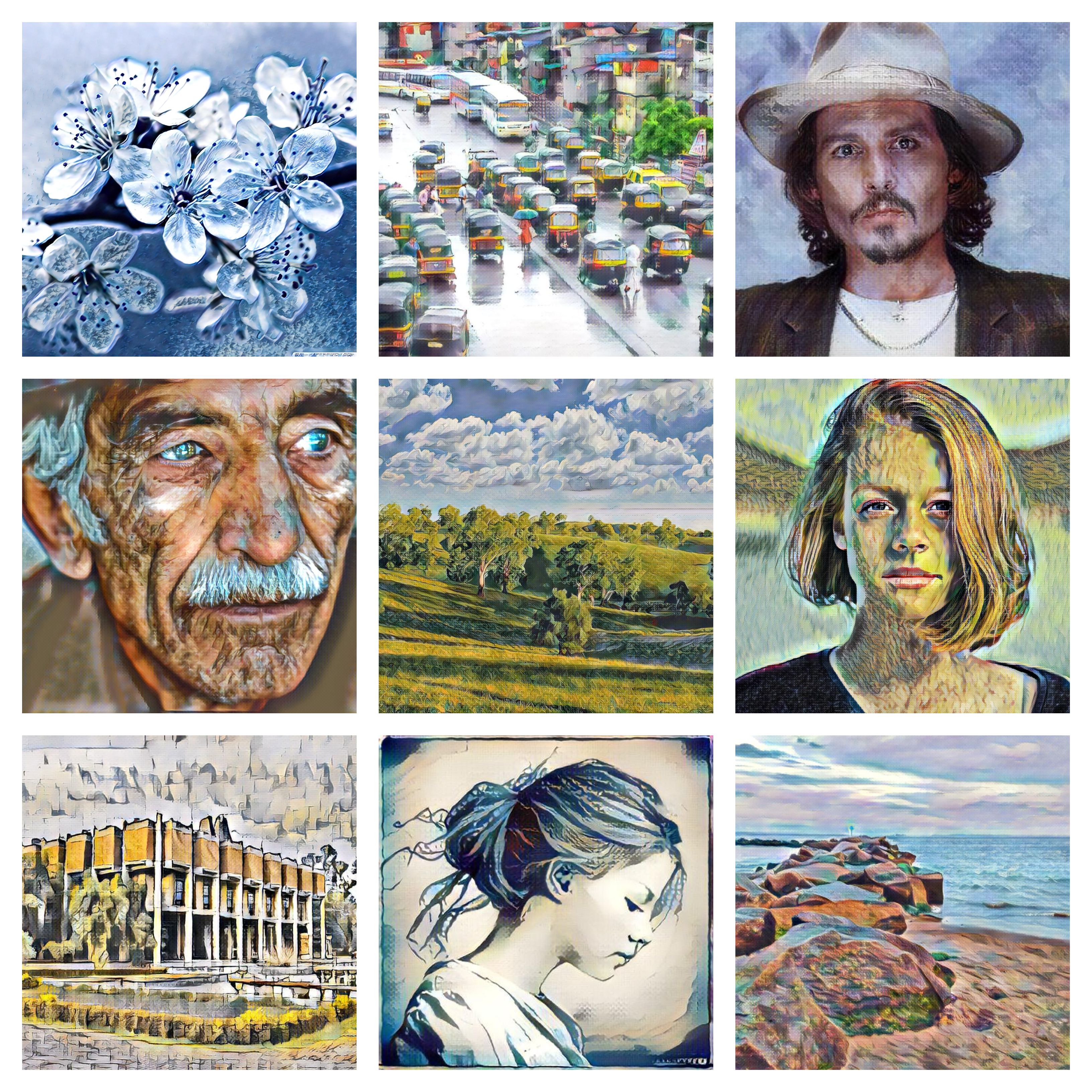}
    \vspace{-6pt}
    \caption{\textbf{Stylization results}. A few stylization results generated using \modelname \xspace with our base network.}
    \label{fig:attendIN-ex}
    \vspace{-6pt}
\end{figure}

Despite much progress being made on normalization-based techniques, one remaining challenge lies in the lack of ability to transfer different textures to different spatial locations of the image with semantic awareness. Although spatially adaptive normalization has been recently used to good effect in other tasks \cite{wang2020neural,park2019semantic}, similar ideas have not been explored in style transfer yet.
\par Attention-based models \cite{park2019arbitrary} \cite{yao2019attention} have shown another successful stream in style transfer. They attempt to resolve this limitation by employing self-attention over the spatial dimensions when transforming the content features. However, these sometimes cause unwanted patterns in output
(Fig. \ref{fig:eyes}).

\par Motivated by these observations, this paper aims to combine the best of both worlds: self-attention and normalization. Our underlying motivation is to make normalization more spatially flexible and semantically aware. We achieve this by using a self-attention based module to learn to generate the parameters of our spatially-adaptive normalization module. We further introduce a new factorized parameter structure to the normalization layer. The resulting new module, called Self-Attentive Factorized Instance Normalization (\textbf{\modelname}), has a plug-and-play nature and can boost other state-of-the-art normalization-based style transfer methods. In addition, we 
develop a novel base network composed of Wavelet Transform for multi-scale style transfer. When combined with \modelname, visually appealing stylization results can be obtained without additional artifacts and distortions.

\section{Related Work}
\subsection{Normalization in NST}
The essential task in Neural Style Transfer (NST) is the transformation of content-image features in a way that their style characteristics match those of the style image. 
\par In optimization based methods (i.e. \textit{single-style-per-model}) \cite{ulyanov2016texture} \cite{johnson2016perceptual}, the use of batch-normalization (BN) \cite{ioffe2015batch} was purely for it's standard purpose of easing neural network training. The work by \cite{ulyanov2016instance} introduced instance-normalization (IN) and its advantage over batch normalization due to it's ability to discard instance-specific contrast information. With new models performing \textit{multiple-styles-per-model} or \textit{arbitrary-styles-per-model}, normalization
(i.e affine shifting and scaling of mean-variance-normalized content features) became a way of imbuing the content features with the style characteristics of the input style. For instance, distinct scaling/shifting parameters are learned for each new style during training in Conditional-IN \cite{dumoulin2016learned}, thus allowing the model to achieve stylization for multiple styles.

\par The main task in accomplishing \textit{arbitrary} style transfer using the normalization based approach is to compute the normalization parameters at test time. AdaIN \cite{huang2017arbitrary} showed that even parameters as simple as the channel-wise mean and variance of the style-image features could be effective. Batch-IN \cite{nam2018batch} used a combination of BN and IN whereas Dynamic-IN \cite{jing2019dynamic} used a learned convolutional network to extract the scaling and shifting parameters given the style and content images.

\begin{figure}[h!]
    \centering
    \includegraphics[scale=0.148]{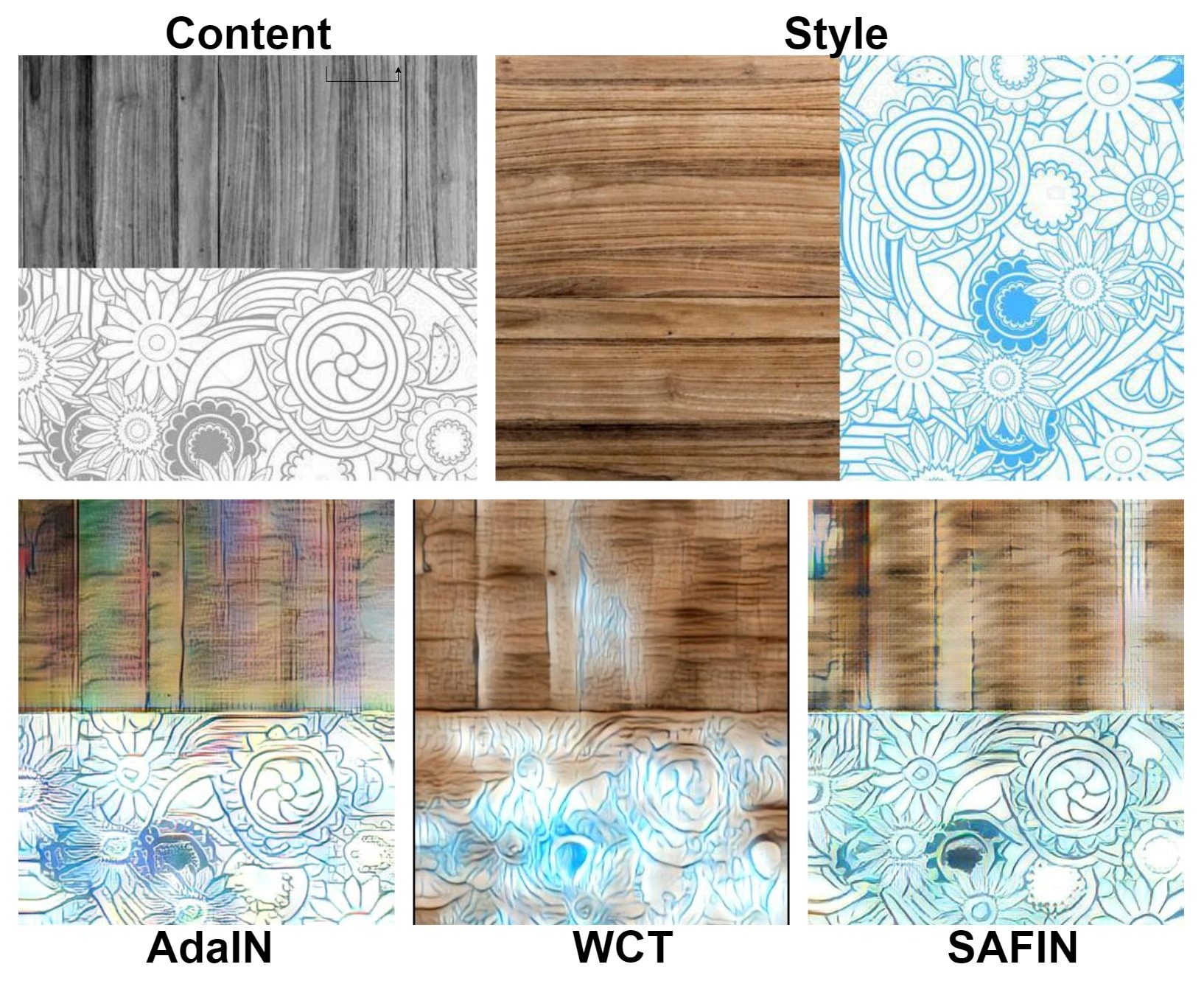}
    \caption{\textbf{Semantic awareness}. Attention-based models show some capability of matching patterns to decide what texture to transfer to different spatial locations. \modelname \xspace has no obvious spillover, whereas WCT and AdaIN suffer from spillovers.}
    \label{fig:semantic}
\end{figure}

\par A major limitation of the current normalization based approaches is their inability to transfer diverse and fine stroke patterns from different parts of the style image to the semantically similar regions in the content image (Fig \ref{fig:semantic}). 
This is because the normalization is either not spatially adaptive \cite{huang2017arbitrary} \cite{nam2018batch} or lacks an appropriate mechanism to infer and transfer the diversity in style patterns onto the normalization parameters.

\subsection{Other Techniques}
WCT \cite{li2017universal} performs the transformation on the content features by using the whitening and coloring transform. 
Not only does WCT suffer from transfer of global style characteristics, leading to less varied stroke patterns, but it is also computationally expensive and especially slow and memory consuming for high-resolution images.
\par Initial attempts to solve the problem of transferring different style patterns used a patch-based approach by matching local statistics between semantically matching patches \cite{shih2014style} \cite{liao2017visual}. Though these methods achieve visually pleasing results, their application is limited to content-style combinations with dense semantic correspondence.

\subsection{Attention in NST}
\begin{figure}[t!]
    \centering
    \includegraphics[scale=0.27]{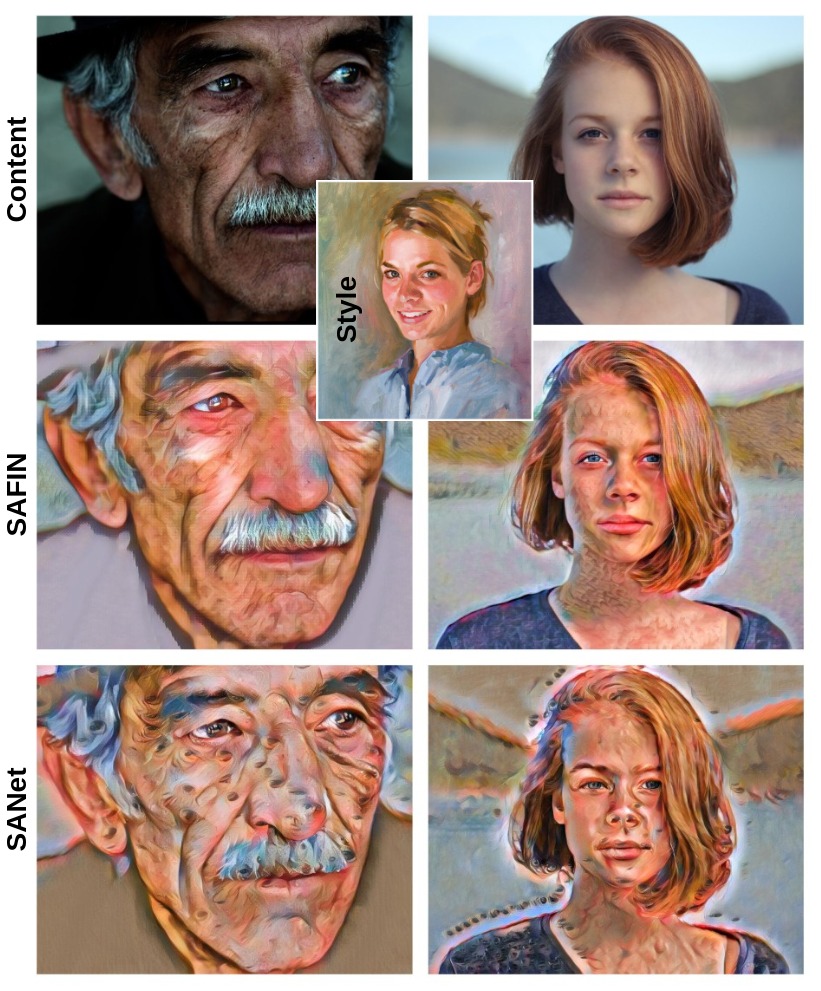}
    \vspace{-6pt}
    \caption{\textbf{Extraneous artifacts}. SANet produces eye-like artifacts (zoom in for greater clarity) due to the fine-grained nature of its attention target while \modelname \xspace does not.}
    \label{fig:eyes}
    \vspace{-6pt}
\end{figure}

Self-Attention \cite{vaswani2017attention} has been applied to the problem of style transfer in SANet \cite{park2019arbitrary} since it suggests a potential way to designate different style patterns to different regions in an image. 
However, in SANet, the fine-grained nature of the attention target and direct incorporation within the content feature maps without further processing sometimes leads to extraneous artifacts and distortions (Fig. \ref{fig:eyes}). 
\par Our method builds upon the ability of SANet and attempts to avert its drawbacks by using it in a novel manner. We use the self-attention module as a part of a convolutional network trained to infer the normalization parameters. This is in contrast to SANet, which directly uses the attention output to infer the feature map for the stylized output. 

\section{Approach}
\begin{figure*}[t]
\centering
\includegraphics[width=17.5cm, height=6cm]{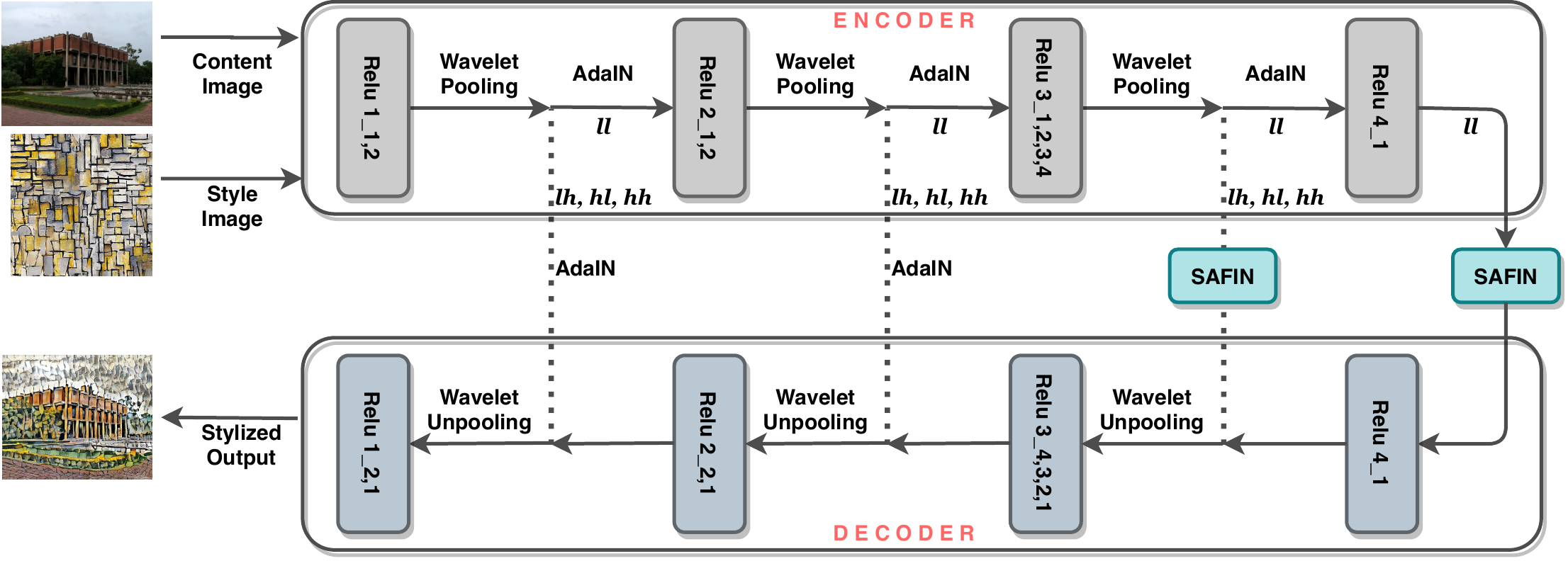}
\caption{\textbf{Network Architecture}. We train a multi-scale encoder-decoder network with two SAFIN modules and Wavelet Transform to stylize the low frequency component (LL, shown through bold lines) as well as the high frequency components (LH, HL, and HH, shown through dotted lines). AdaIN \cite{huang2017arbitrary} is used at different scales to further enrich the stylization.} 
\label{fig:arch}
\vspace{-6pt}
\end{figure*}

\subsection{Factorized Instance Normalization (FIN)}
We present a novel normalization module called Factorized Instance Normalization (FIN). Given the content feature maps $F_c \in \mathbb{R}^{H * W * C}$ we transform the style characteristics of these features to match the style characteristics of the style image while maintaining it's semantic information. We achieve this via the following transformation:
\begin{align} \label{eqn:FIN}
    \text{FIN}(F_c) &= \gamma_s \odot \big(\gamma_{ind} \odot \overline{F_c} + \beta_{ind} \big) + \beta_s
\end{align}
where $\overline{F_c} \in \mathbb{R}^{H * W * C}$ is the normalized content feature obtained by applying channel-wise instance normalization on $F_c$. $\odot$ denotes element-wise multiplication.

For the shifting and scaling operations (Eq. \ref{eqn:FIN}), we use two types of learned normalization parameters:
\begin{enumerate}
    \item $\gamma_{ind}$ and $\beta_{ind}$ represent the \textit{style independent} parameters and are single dimension vectors along the channel dimension (i.e. $\gamma_{ind}, \beta_{ind} \in \mathbb{R}^{C}$). These parameters are meant to capture the shared transformation between the different style images, by transforming $\overline{F_c}$  before the style dependent transformation.
    
    \item  On the other hand, $\gamma_s$ and $\beta_s$ are inferred from the input style image. $\gamma_s$, $\beta_s$ are 3 dimensional; along the channel dimension as well as the spatial dimensions (i.e. $\gamma_s, \beta_s \in \mathbb{R}^{H * W* C}$). Thus, element-wise shifting and scaling of the content features using $\gamma_s$ and $\beta_s$ allows for greater flexibility in transferring diverse textures to different spatial regions on the content image. However, to generate rich stylization, it is not enough to simply use spatially adaptive normalization, since a mechanism is required that can adequately learn to derive these normalization-parameters from the input style image. Thus, $\gamma_s$ and $\beta_s$ are inferred from the style image using a network employing self-attention followed by convolution. This mechanism is explained in subsection \ref{subsec:SA}.
\end{enumerate}

\subsection{Self-Attention Mechanism (SA)}
\label{subsec:SA}
We employ Self-Attention \cite{vaswani2017attention} to capture the semantic correspondence between the content feature maps $F_c$ and the style feature maps $F_s$ (extracted from the encoder) for generating the style-based normalization parameters $\gamma_s$ and $\beta_s$.

\abovedisplayskip=0pt
\begin{gather} \label{eqn:gammabeta}
    \gamma_s = \Call{ReLU}{W_{\gamma} \otimes \text{SA}(\overline{F_c}, \overline{F_s})} \\
    \beta_s = \Call{ReLU}{W_{\beta} \otimes \text{SA}(\overline{F_c}, \overline{F_s})}
\end{gather}

Here $\overline{F_c}, \overline{F_s} \in \mathbb{R}^{H * W * C}$ are channel-wise instance normalized content and style feature maps respectively. $W_{\gamma}$, $W_{\beta}$ represent $1 * 1$ convolutional layers and $\otimes$ denotes the convolution operation. Our Self-Attention module (SA) is similar to that used by \cite{park2019arbitrary}, though it is employed for a different purpose. \cite{park2019arbitrary} used self-attention to directly modify the content feature maps, whereas we instead use it to generate the style-dependent normalization parameters of our FIN module.
\subsection{Wavelet Transform}
For better preservation of content features in the stylized outputs, we use Wavelet Pooling (to replace Max Pooling) and Wavelet Unpooling (to replace Upsampling) \cite{guf1996haar}, a pooling technique which allows exact reconstruction due to it's invertible nature.
This technique has been previously employed in photo-realistic style transfer \cite{yoo2019photorealistic}, but not, to the best of our knowledge, in artistic style transfer.
\par Wavelet pooling uses four distinct pooling kernels  $\{LL^\intercal,LH^\intercal,HL^\intercal,HH^\intercal\}$ at every pooling layer, made by combining the high-pass filter $H^\intercal = \frac{1}{\sqrt{2}}[ \hspace{1mm}-1 \hspace{3mm} 1 \hspace{1mm}]$ and the low-pass filter $L^\intercal = \frac{1}{\sqrt{2}}[ \hspace{1mm}1 \hspace{3mm} 1 \hspace{1mm}]$ 
. We use the terms LL, LH, HL, and HH to refer to the output of the respective kernel.

\subsection{Network Architecture}
Figure \ref{fig:arch} shows the network architecture, which follows the commonly employed encoder-decoder structure \cite{huang2017arbitrary}, \cite{li2017universal}, \cite{park2019arbitrary}. The encoder has the same structure as the VGG-19 network, and the decoder is symmetric to the encoder.

The loss function for training the \modelname \xspace module and the decoder is given as: $L = L_{c} + \lambda_{s}*L_{s}$, where $L_{c}$ and $L_{s}$ are content and style losses and $\lambda_{s}$ is the weight assigned to the style loss relative to the content loss. 

The content loss is the Euclidean distance between LL components of stylised output image encoding $\mathcal{E}(I_{cs})$ and content image encoding $\mathcal{E}(I_{c})$ extracted from the layer ReLU\_4\_1:
$L_{c} = ||\mathcal{E}_{LL}^{4.1}(I_{cs}) - \mathcal{E}_{LL}^{4.1}(I_{c})||_{2}$ 

The style loss is the Euclidean distance between mean and variance of LL components of stylised output image encoding $\mathcal{E}(I_{cs})$ and style image encoding $\mathcal{E}(I_{s})$ from the layers ReLU\_X\_1 where $\text{X} \in \{1, 2, 3, 4\}$: 
\begin{align*}
    L_{s} =& \sum\limits_{k=1}^4 ||\mu(\mathcal{E}_{LL}^{k.1}(I_{cs})) - \mu(\mathcal{E}_{LL}^{k.1}(I_{s}))||_{2} + \\
    & \hspace{3cm} ||\sigma(\mathcal{E}_{LL}^{k.1}(I_{cs})) - \sigma(\mathcal{E}_{LL}^{k.1}(I_{s}))||_{2}
\end{align*}

\par Further details of the network architecture are described in the appendix.
\vspace{-3pt}
\section{Experiments}
We trained our network using MS-COCO dataset \cite{lin2014microsoft} for content images and WikiArt dataset \cite{phillips2011wiki} for style images. The detailed experimental settings are included in the appendix. Code and pretrained model are available here \url{https://github.com/Aaditya-Singh/SAFIN}.

\subsection{Generalizability}
\begin{figure}[t!]
\centering
   \begin{subfigure}{0.43\textwidth}
   \includegraphics[width=1\linewidth,height=12cm]{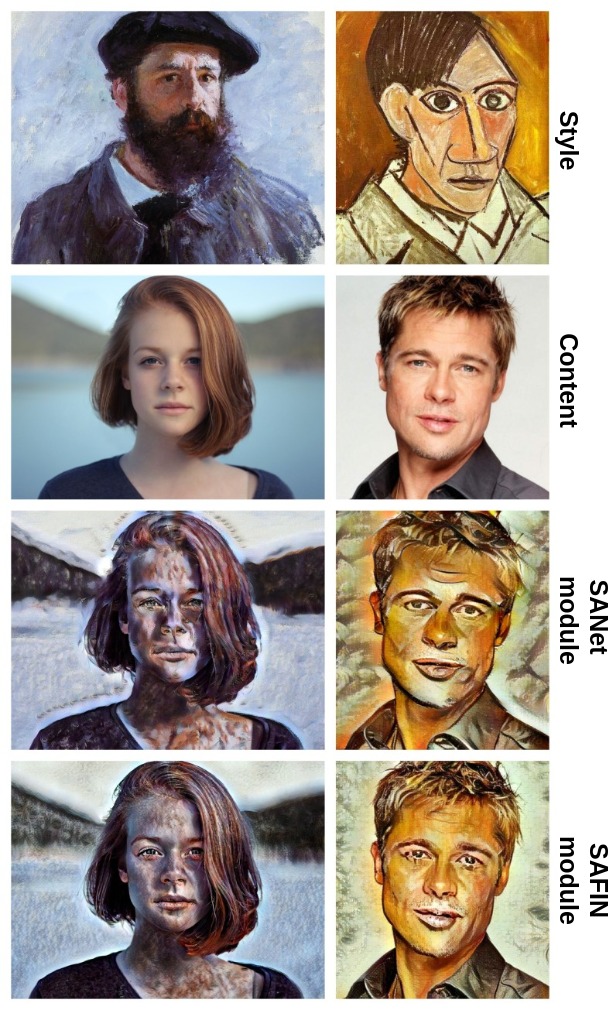}
   \caption{\modelname \xspace module when substituted in place of the SANet module in SANet's base network, leads to reduced unwanted distortions.}
   \label{fig:safin-v-sanet-img} 
    \end{subfigure}
    \begin{subfigure}{0.48\textwidth}
   \includegraphics[width=1\linewidth]{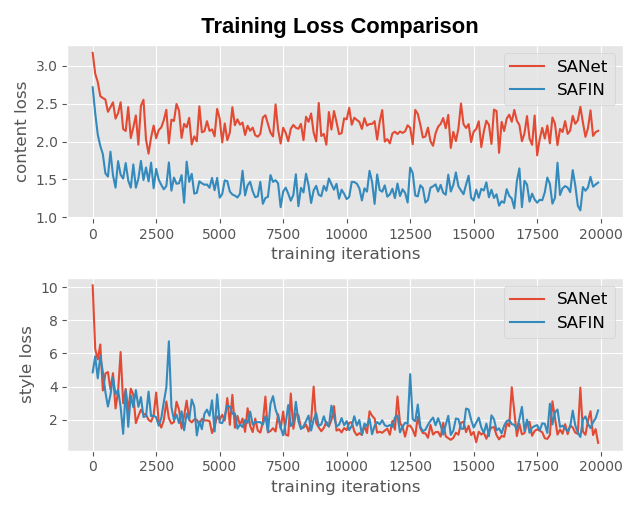}
   \caption{Training loss comparison: SANet's base network with SANet's style transfer module vs \modelname \xspace module.} 
   \label{fig:safin-v-sanet-loss}
    \end{subfigure}
\caption{\textbf{\modelname \xspace (with SANet's base network) vs SANet.} }
\vspace{-6pt}
\end{figure}

To demonstrate the superiority as well as flexibility of our spatially adaptive normalization technique, we show how \modelname \xspace module when substituted in to replace the SANet \cite{park2019arbitrary} module within the SANet base network, also mitigates the extraneous artifacts to some extent (Fig. \ref{fig:safin-v-sanet-img}). The training-loss comparison plot (Fig. \ref{fig:safin-v-sanet-loss}) shows a lower content loss with the \modelname \xspace module compared to the SANet module thus demonstrating that the usage of \modelname \xspace results in better content preservation. Style losses are similar. Indeed, our module when inserted into SANet's base network provides a comparable quality of stylization along with reduced unwanted distortions.

\subsection{Ablation Study}
\subsubsection{Effect of removal of Self-Attention from \modelname}
\begin{figure}[t!]
    \centering
    \includegraphics[scale=0.27]{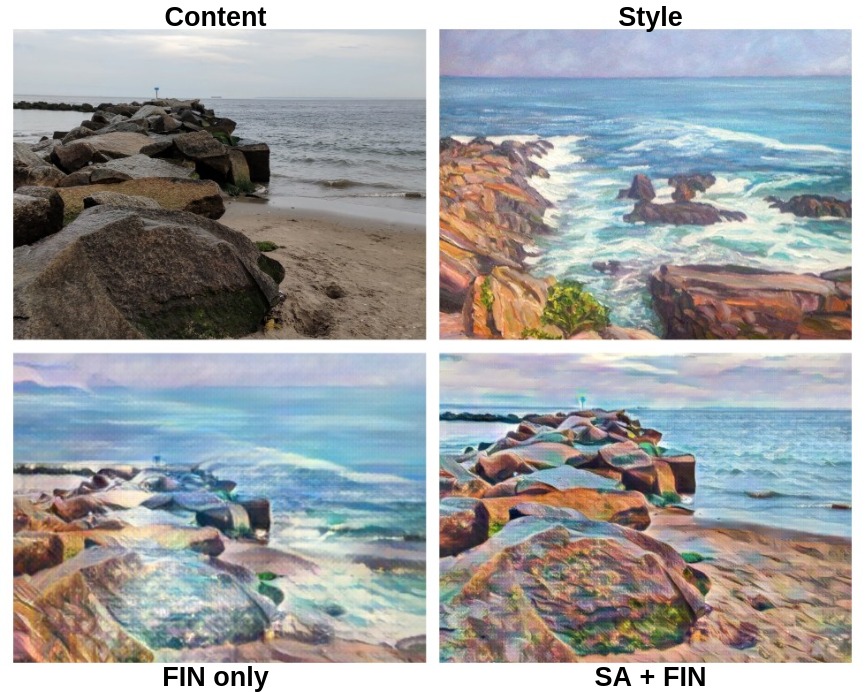}
    \caption{\textbf{Effect of removal of Self-Attention (SA) module}. FIN transfers the general color and texture patterns but causes spillovers, whereas \modelname \xspace produces a more semantically coherent stylization.}
    \label{fig:fin_vs_safin}
    \vspace{-6pt}
\end{figure}

In order to understand the effectiveness of Self-Attention (SA) mechanism within the overall \modelname \xspace module, we compare the results when \modelname \xspace module is replaced with FIN module (normalization without self-attention) in our network. From figure \ref{fig:fin_vs_safin}, it can be observed that while FIN module enables the transfer of general color and texture from the style image, self-attention allows the network to transfer the style characteristics from semantically similar regions in the style image leading to an enhanced stylization.
\par Further ablation studies and more stylization outputs for comparison are included in the appendix.

\subsection{Quantitative Analysis}
\vspace{-6pt}
\begin{figure}[ht!]
    \centering
    \includegraphics[width=0.98\linewidth]{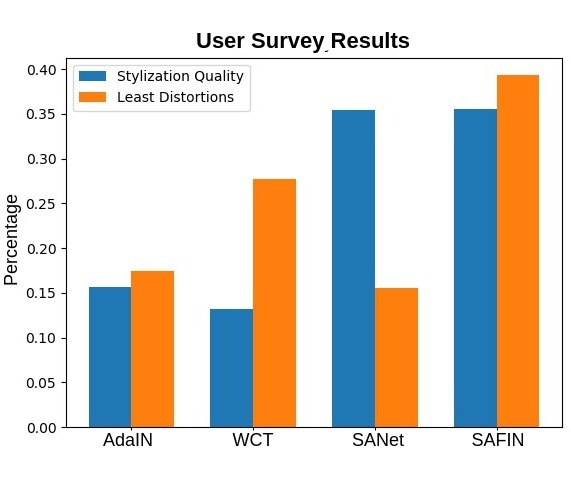}
    \vspace{-10pt}
    \caption{\textbf{User Survey Results}. \modelname \xspace with our base network causes the least amount of artifacts and distortions without compromising the stylization quality.}
    \label{fig:survey}
    \vspace{-6pt}
\end{figure}
We also conducted an extensive survey for a thorough quantitative analysis. There were 25 participants and 40 sets of content-style image pairs along with their stylization outputs using 4 different methods - \modelname \xspace (with the base network), SANet, WCT and AdaIN. For every set, each participant was asked two questions: 
\begin{itemize}[noitemsep]
    \item To choose the best stylization, i.e. which method leads to the best color and texture transfer.
    \item To choose which stylization caused the least amount of unwanted artifacts and distortions.
\end{itemize}
The results (Fig. \ref{fig:survey}) show that our method is preferred by users against other methods, providing better or comparable stylization without leading to additional distortions.

\section{Conclusion}
\vspace{-5pt}
In this paper, we propose a novel module called SAFIN which combines normalization and self-attention in a unique way to mitigate the issues with previous style transfer techniques. Extensive experiments demonstrate its effectiveness. We show the generalization capability of the proposed module by showing how it improves the stylization and losses of another state-of-the-art method when incorporated within its base network. We also develop a novel base network composed of Wavelet Transform for multi-scale style transfer that helps in preserving more content information. Experimental results illustrate the effectiveness of the proposed method.

\bibliographystyle{IEEEbib}
\bibliography{icme2021template}

\end{document}